\documentclass[conference]{IEEEtran}
\IEEEoverridecommandlockouts
\usepackage{cite}
\usepackage{amsmath,amssymb,amsfonts}
\usepackage{algorithmic}
\usepackage{graphicx}
\usepackage{textcomp}
\usepackage{xcolor}
\usepackage{hyperref}
\usepackage{fixfoot}
\usepackage{etoolbox}
\usepackage{graphicx}
\usepackage{multirow}
\def\BibTeX{{\rm B\kern-.05em{\sc i\kern-.025em b}\kern-.08em
    T\kern-.1667em\lower.7ex\hbox{E}\kern-.125emX}}
    
\DeclareFixedFootnote{\reconstructeddataset}{\url{https://github.com/sweta20/Detecting-Cyberbullying-Across-SMPs}}
    
\begin{document}

\title{Generalisation of Cyberbullying Detection\\
\thanks{This research was made possible by the financial, material, and technical support of Two Hat Security Research Corp., and the financial support of the Canadian research organization MITACS.}
}

\author{\IEEEauthorblockN{Marc-André Larochelle}
\IEEEauthorblockA{\textit{Computer Science and Software Engineering Department} \\
\textit{Université Laval}\\
Québec, Canada \\
marc-andre.larochelle.1@ulaval.ca}
\and
\IEEEauthorblockN{Richard Khoury}
\IEEEauthorblockA{\textit{Computer Science and Software Engineering Department} \\
\textit{Université Laval}\\
Québec, Canada \\
richard.khoury@ift.ulaval.ca}
}

\maketitle

\begin{abstract}

Cyberbullying is a problem in today's ubiquitous online communities. Filtering it out of online conversations has proven a challenge, and efforts have led to the creation of many different datasets, all offered as resources to train classifiers. Through these datasets, we will explore the variety of definitions of cyberbullying behaviors and the impact of these differences on the portability of one classifier to another community. By analyzing the similarities between datasets, we also gain insight on the generalization power of the classifiers trained from them. A study of ensemble models combining these classifiers will help us understand how they interact with each other. 
\end{abstract}


\section{Introduction}
Online social interactions are commonplace today. Unfortunately, with the positive benefits of bringing together people in an unprecedented way, has come the negative impacts, namely the spread of cyberbullying and its social toll. For example, a recent study in Canada \cite{hango2016cyberbullying} found that 17\% of 15-to-29-year-old Internet users (or 1.1 million individuals) had experienced cyberbullying, that the problem disproportionately affected women (19\%), low-income people (24\%), and homosexuals (34\%), and that 20\% of victims developed emotional, psychological or mental health conditions as a result.


Although many solutions to this problem have been proposed in the literature and in industry, none of them agree on precisely what constitutes cyberbullying. For example, \cite{van2018automatic} has found that cyberbullying can include general insults, cursing, sexism or hate speech, defamation, sexual harassment, threats or blackmail, and even polite support of the aggressor, directed against a victim or their friends and family, which may occur once or be repeated, which may or may not involve a power imbalance between the aggressor and victim, and which may be an intentional attack or a misunderstanding. 





In this paper, we aim to explore the relationship and distinctions that exist between datasets that implement different definitions of cyberbullying, the possibilities and limitations to applying a system trained on one dataset to filter cyberbullying in a new context, and to discover empirically the best way to combine them to create a generalized cyberbullying detection system. In line with these objectives, section \ref{sec:datasets} will present a selection of datasets that each cover a subspace of the general space of cyberbullying. In section \ref{sec:similarity}, we study the similarities in the language used in these datasets. Next, in section \ref{sec:generalization}, we train a deep-learning model on each dataset in order to study their generalization power, before exploring different ways to combine them into a general cyberbullying detector in section \ref{sec:ensemble}. In all cases, the results we obtain are thoroughly analyzed. Finally, section \ref{sec:conclusion} will draw some concluding remarks.


\section{Datasets}\label{sec:datasets}



For our research, we have selected eight datasets pertaining to various types of behaviors that fit the general definition of cyberbullying. Many of these datasets have labels to designate specific types of behaviors which are either not found in other datasets or defined in a different manner. In order to make the datasets directly comparable, we merged all these labels into a positive "cyberbullying" class, opposing the much more common negative class of messages that are not cyberbullying. In addition, some of these datasets came divided into training and testing sets. For the others, we randomly divided the datasets into a 20\% test set and 80\% training set, which we further divided into 80\% training and 20\% validation sets. Basic statistics for all the datasets are given in table \ref{tbl:wordcounts}.


\subsection{Hate Speech and Offensive Language}

This dataset\footnote{\url{https://github.com/t-davidson/hate-speech-and-offensive-language}} was collected by \cite{hateoffensive} by searching Twitter for tweets containing hate speech terms from the lexicon \textit{Hatebase.org}. Out of the 85.4 million tweets gathered, 25,000 tweets were randomly selected and annotated by three or more annotator to one of three labels: if it contains hate speech, if it features offensive language without hate speech, or if it contains neither hate speech nor offensive language. A tweet's final label was the majority decision, and tweets for which no majority decision existed were filtered out. This gave them a corpus of 4,163 tweets that do not contain hate speech nor offensive language, 19,190 that contain offensive language, and 1,430 that are considered hate speech, making it the only corpus in our study imbalanced in favor of the positive class.

\subsection{Racism and Sexism}
The authors of \cite{waseem-hovy:2016:N16-2} designed a list of slurs and terms identifying religious, sexual, gender, and ethnic minorities. They then sampled Twitter for tweets using these words, and refined their list based on the results. They manually annotated each sampled tweet as sexist, racist, or neither, and had an expert review their annotations in order to mitigate possible bias. We used a version of their dataset\reconstructeddataset made available by \cite{deep_learning_across_smp}. 

\subsection{Bullying}





This dataset was gathered by the authors of \cite{6147681} from the question-answering (QA) platform \textit{Formspring}. They randomly selected 50 users from the 18,554 users of the platform, each with between 1 and 1,000 QA pairs on the site. Each pair was labeled by three annotators who were asked to judge if it contains cyberbullying and, if so, to identify which words or phrases were the reason. This created a labeled dataset of 12,773 QA pairs\footnote{\url{www.kaggle.com/swetaagrawal/formspring-data-for-cyberbullying-detection}}. We follow the work of \cite{deep_learning_across_smp} and concatenate the question and answer into as single message. 

\subsection{Insults in social commentary}


This Kaggle competition dataset \cite{noauthor_detecting_nodate} was gathered from an unspecified social networking site, and the comments it is comprised of were labeled as insulting or not. However, this dataset applies a narrower definition and only labels positive messages if they are obviously insulting to a specific member of the on-going conversation. Messages that insult celebrities and public figures, messages that include insults and racial slurs not directed at a specific person, and subtle insults, are all counted as negative-class messages. 

\subsection{Hate Speech}


This dataset is composed of messages scraped from the white-supremacist internet forum \textit{Stormfront} by the authors of \cite{de_gibert_hate_2018}. The messages were labeled into one of four classes. The "hate" class is for messages that are (i) deliberate attack (ii) directed towards a specific group of people (iii) and motivated by aspects of the group's identity. This definition is a different from traditional hate speech: for example using a racial slur in an offhand manner and without deliberately attacking that racial group is not considered a hate message. The "relation" class is for messages that do not fit in the hate class by themselves, but do when read within the context of the conversation in which they appear. The "skip" class contains non-English messages and gibberish. Finally, the "non-hate" class is for all messages that do not fit the other three categories. The original dataset\footnote{\url{https://github.com/aitor-garcia-p/hate-speech-dataset}} is composed of 9,507 non-hate messages, 1,196 hate messages, 168 relation messages, and 73 skip messages. We redefined the hate class as our positive class and the others as our negative class. The rationale for including relation in the negative class is that most cyberbullying detection systems work on a line-by-line basis and will mark such messages without context as acceptable. 

\subsection{Toxic comment classification}

The second Kaggle dataset \cite{noauthor_toxic_nodate} was gathered from Wikipedia talk pages comments. Each comment was annotated with six different labels: toxic, severely toxic, obscene, threat, insult, and identity hate; however the annotation procedure is not described. Unlike the previous datasets, each comment can be tagged with multiple labels. The dataset contains 21,384 comments labeled as toxic, 1,962 as severely toxic, 12,140 as obscene, 689 as threats, 11,304 as insults, and 2,117 as identity hate. For our work, we assign a comment to our positive class if it contains any one or more of these labels. 

\subsection{Unintended bias in toxicity classification}




This dataset, also from Kaggle \cite{noauthor_jigsaw_nodate}, was obtained from a news website comment filter system called \textit{Civil Comments}. Its 1,999,514 comments were labeled by three to ten crowd-sourced annotators on average (and sometimes up to a thousand annotators) into six labels: severe toxicity, obscene, threat, insult, identity attack, and sexual explicit. Any and all labels chosen by half or more of annotators was applied to the comment. Once again, we consider a comment as belonging to the positive class if any of the labels was applied to it. 


\subsection{Personal attacks and harassment}





The final dataset we selected for our study was also constructed from Wikipedia talk page comments. The authors of \cite{wulczyn_ex_2017} randomly sampled the 63 million talk-page comments posted between 2004 and 2015. They added comments from a set of users blocked for violating Wikipedia's policy on personal attacks, sampling five comments per user shortly before they were blocked. These were given to a group of annotators, who were asked if each comment contains a personal attack or harassment, whether it is targeted at the recipient or a third party, if it is being reported or quoted, and if it is another kind of attack or harassment. Data quality was assured by requiring that each annotator label ten test comments, and quality control comments were also inserted during the annotation process. Each comment was annotated by ten separate annotators. This resulted in a corpus\footnote{\url{https://doi.org/10.6084/m9.figshare.4054689.v6}} of 115,859 comments, of which 13,590 were found to contain personal attack or harassment. We should note that there is an overlap between the training set of the toxic comment classification competition and test set of this dataset. After the competition launched, its organizers realized that some of their test corpus was included in this dataset\footnote{\url{https://www.kaggle.com/c/jigsaw-toxic-comment-classification-challenge/discussion/46177}}. They corrected the problem by moving these messages to their training corpus, which solved the issue of getting a high score in the competition through overfitting, but left the overlap in place. 

\section{Dataset vocabulary comparison}\label{sec:similarity}
The eight datasets we selected were collected from platforms with different messaging formats, from limited-length tweets to question-answer pairs to forum conversations. They all pertain to some aspects of cyberbullying \cite{van2018automatic}, although only \cite{6147681} explicitly names it as such and the problematic behaviors they monitor varies greatly, from the focused scope of the Twitter corpora to the wide range of behaviors labeled in the Kaggle datasets. Some behaviors are common; hate speech is explicitly labelled in five of the eight datasets. And some behaviors are unique to some corpora; threats are explicitly noted in only two corpora and sexually-explicit comments in one. This wide variety of forms of cyberbullying and its consequences are the main focus of our study.

To begin, we will study the impact of this diversity on the vocabulary of the datasets using both traditional cosine similarity and word embeddings First, we divide each dataset into its positive and negative portions and treat each as a separate dataset. Basic statistics are given in Table \ref{tbl:wordcounts}. In this table, the dataset letters refer to the subsections of Section \ref{sec:datasets}, and the subscript + and - to the positive and negative class.

\begin{table}
\caption{Number of messages, unique words, and total word count in each dataset}
\begin{center}
\begin{tabular}{|l|r|r|r|}
\hline
\textbf{Dataset} & \textbf{Messages}  & \textbf{Unique words} & \textbf{Total words} \\
\hline
A$_-$ & 3,002 & 10,989 & 34,606\\
A$_+$ & 14,841 & 23,193 & 155,359\\
B$_-$ & 7,957 & 15,153 & 73,973\\
B$_+$ & 3,627 & 9,531 & 40,027\\
C$_-$ & 8,619 & 14,445 & 104,946\\
C$_+$ & 556 & 2,554 & 8,049\\
D$_-$ & 2,898 & 13,195 & 56,848\\
D$_+$ & 1,049 & 4,377 & 13,183\\
E$_-$ & 7,002 & 12,468 & 61,023\\
E$_+$ & 877 & 3,706 & 10,273\\
F$_-$ & 114,677 & 162,524 & 4,410,757\\
F$_+$ & 12,979 & 32,001 & 398,342\\
G$_-$ & 1,358,749 & 284,164 & 38,232,678\\
G$_+$ & 85,150 & 63,441 & 2,032,318\\
H$_-$ & 49,155 & 99,083 & 1,983,200\\
H$_+$ & 6,463 & 21,748 & 235,820\\
\hline
\end{tabular}
\label{tbl:wordcounts}
\end{center}
\end{table} 

To compute cosine similarity, we convert each dataset into its bag-of-word representation and compute the TFIDF value of each word using the standard formula:
\begin{equation}
tfidf(w, d) = (1 + \log n_{w,d}) \times  \log\frac{D}{D_w} \label{tfidf}
\end{equation}
\noindent where the tfidf value of word $w$ in dataset $d$ is the log normalization of the number of times the word occurs in the dataset ($n_{w,d}$) times the inverse log of the number of datasets $D$ (16 since we handle the positive and negative class of each dataset separately here) and $D_w$ the number of datasets containing word $w$. Using the log normalization of the word count (defined as 0 if a word does not occur in a corpus) instead of using the word count directly helps mitigate the impact of the extreme difference in the size of our datasets shown in table \ref{tbl:wordcounts}, by focusing on the order of magnitude of the counts instead of their values. Once each dataset is represented by its TFIDF-weighted word vector, we compute the cosine similarity between each pair of dataset:

\begin{equation}
\cos ({\bf A},{\bf B}) = \frac{{\bf A} \cdot {\bf B}} {\|{\bf A}\| \cdot \|{\bf B}\|}
\label{cosine_distance}
\end{equation}

\noindent where \textit{\bf A} and \textit{\bf B} are the word vectors of two datasets. The cosine similarity is a vector distance metric in the vocabulary space, ranging from 0 for two completely different vectors to 1 for two identical vectors. Measuring this for each pair of our 16 classes gives the results presented in table \ref{tab:cosine_similarity}. These results show that, for 6 of the 8 datasets, the most similar dataset to its positive class is its negative class and vice-versa. The two exceptions are datasets F and H, where the negative and positive class of each is twice as similar to the negative and positive class of the other than to its own complementary class. Since both of these datasets were constructed from Wikipedia talk page comments, this similarity is not surprising. The similarity between the positive and negative class of dataset G is a lot higher than the similarity between the classes of other datasets, but this can be explained by the fact this dataset comes from news comments and so its conversations were constrained to news topics and vocabulary, unlike the other corpora in which users could discuss any topic at all.




In fact, what is surprising in table \ref{sec:similarity} is not the similarities we find, but how few of them we find. Datasets A, B, C, D and E have nearly no similarity to any other dataset. Datasets A and B, both from Twitter, have no more similarity to each other than they do to datasets from other sources. Likewise, the positive classes of datasets A and E, both focusing on hate speech, have no more in common than they do to other datasets, positive or negative. Even the similarity between the positive and negative class of each dataset is rather low. Normally we would expect language to be homogeneous throughout a dataset and to vary mainly on the presence or absence of class-specific cyberbullying keywords, and since those would be a minority of the words used the similarity between the positive and negative classes should be high. The low values observed indicate the opposite, that the positive and negative classes of each dataset differ also on the non-cyberbullying vocabulary used. 
Finally, the low similarity between the positive classes of different datasets shows that the vocabulary marking cyberbullying is very different from one dataset to the next. This can be attributed in part to the different cyberbullying behaviors each dataset measures and to the wildly different platforms each dataset was collected from, and of course to the diversity and flexibility of the English language. The consequence, however, is that we should expect a system trained to recognize cyberbullying in one dataset to have a lot of difficulty picking out cyberbullying in another.


\begin{table*}[htpb]
\caption{Cosine Similarity between each pair of datasets}
\begin{center}
\begin{tabular}{|c|c|c|c|c|c|c|c|c|c|c|c|c|c|c|c|c|}
\hline
 & A$_-$ & A$_+$ & B$_-$ & B$_+$ & C$_-$ & C$_+$ & D$_-$ & D$_+$ & E$_-$ & E$_+$ & F$_-$ & F$_+$ & G$_-$ & G$_+$ & H$_-$ & H$_+$ \\
\hline
A$_-$ & 1.000 &  &  &  &  &  &  &  &  &  &  &  &  &  &  &  \\
A$_+$ & 0.212 & 1.000 &  &  &  &  &  &  &  &  &  &  &  &  &  &  \\
B$_-$ & 0.013 & 0.010 & 1.000 &  &  &  &  &  &  &  &  &  &  &  &  &  \\
B$_+$ & 0.009 & 0.007 & 0.326 & 1.000 &  &  &  &  &  &  &  &  &  &  &  &  \\
C$_-$ & 0.014 & 0.027 & 0.011 & 0.006 & 1.000 &  &  &  &  &  &  &  &  &  &  &  \\
C$_+$ & 0.005 & 0.016 & 0.003 & 0.002 & 0.188 & 1.000 &  &  &  &  &  &  &  &  &  &  \\
D$_-$ & 0.014 & 0.013 & 0.011 & 0.009 & 0.013 & 0.004 & 1.000 &  &  &  &  &  &  &  &  &  \\
D$_+$ & 0.013 & 0.013 & 0.009 & 0.007 & 0.008 & 0.005 & 0.249 & 1.000 &  &  &  &  &  &  &  &  \\
E$_-$ & 0.013 & 0.012 & 0.015 & 0.011 & 0.016 & 0.003 & 0.024 & 0.013 & 1.000 &  &  &  &  &  &  &  \\
E$_+$ & 0.009 & 0.010 & 0.010 & 0.008 & 0.008 & 0.002 & 0.015 & 0.016 & 0.094 & 1.000 &  &  &  &  &  &  \\
F$_-$ & 0.026 & 0.024 & 0.032 & 0.021 & 0.042 & 0.010 & 0.048 & 0.023 & 0.074 & 0.032 & 1.000 &  &  &  &  &  \\
F$_+$ & 0.024 & 0.029 & 0.033 & 0.024 & 0.041 & 0.018 & 0.048 & 0.037 & 0.066 & 0.038 & 0.274 & 1.000 &  &  &  &  \\
G$_-$ & 0.030 & 0.030 & 0.035 & 0.022 & 0.048 & 0.012 & 0.052 & 0.025 & 0.067 & 0.032 & 0.284 & 0.151 & 1.000 &  &  &  \\
G$_+$ & 0.036 & 0.036 & 0.046 & 0.034 & 0.051 & 0.013 & 0.074 & 0.042 & 0.085 & 0.049 & 0.276 & 0.206 & 0.545 & 1.000 &  &  \\
H$_-$ & 0.026 & 0.024 & 0.036 & 0.024 & 0.042 & 0.010 & 0.051 & 0.024 & 0.083 & 0.038 & 0.519 & 0.301 & 0.258 & 0.283 & 1.000 &  \\
H$_+$ & 0.021 & 0.024 & 0.031 & 0.022 & 0.035 & 0.015 & 0.042 & 0.032 & 0.062 & 0.036 & 0.228 & 0.510 & 0.126 & 0.182 & 0.303 & 1.000 \\
\hline
\end{tabular}
\label{tab:cosine_similarity}
\end{center}
\end{table*}

\begin{figure}[htbp]
\centerline{\includegraphics[width=0.5\textwidth]{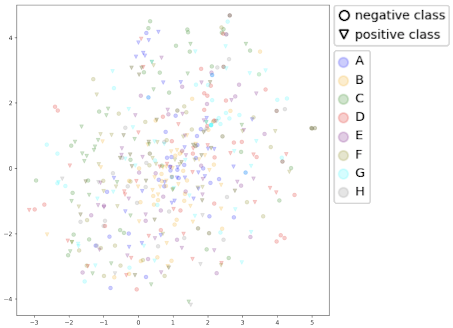}}
\caption{t-SNE projection of the top 30 words of the positive and negative class of each dataset.}
\label{t_sne}
\end{figure}

Next, we select the 30 most important words of each dataset class according to the TFIDF value, and get their word embedding representation using FastText. While cosine similarity considers exact word matches and word counts, word embeddings highlights semantic similarity between different words. We project these 300-dimensions word embeddings onto a 2D map using the t-distributed stochastic neighbor embedding (t-SNE) \cite{t_sne_article}, a nonlinear dimensionality reduction technique which projects high-dimensional objects onto a plane by mapping similar vectors to nearby points and dissimilar vectors to distant points. This results in the graphical representation of figure \ref{t_sne}. The decision to limit ourselves to 30 words per class is simply to avoid overcrowding the graphic. 

This representation of the 480 most significant words of our datasets illustrates the vocabulary diversity problem mentioned earlier. We do observe some small homogeneous regions, mostly on the edges of the figure, where positive or negative words of several datasets cluster near each other. These regions represent words of similar meaning labeled in the same class in different datasets. Ideally we would want the entire graphic to have regions for words of clearly safe meaning and for words of clearly cyberbullying intent agreed upon by all datasets. However, most of the graphic is a large heterogeneous zone, with words of similar meanings labeled in the positive class in one dataset and the negative class in another. These are neutral-meaning words that appear in only one class of a dataset, and thus become significant indicators of that class. Figure \ref{t_sne} shows that this is very frequently the case. Combined with our previous observation on cosine similarity, this further highlights the difficulty of transferring a cyberbullying detector trained on one dataset to another. Not only is there little agreement on how to define cyberbullying and very little vocabulary in common between datasets, but words with similar meanings are labeled in contradictory ways.


\section{Generalization Experiment}\label{sec:generalization}



In this first experiment, we explore the generalization power of a cyberbullying detector trained on any one of the datasets of section \ref{sec:datasets} to the other datasets. This is an important result to document: while every one of the datasets has been used to train detectors that perform very well on a test corpus subset of itself, experiments on other datasets are more rare in the literature, and when done \cite{van2018automatic,emmery2019current,deep_learning_across_smp} they do not explore the limitations of this transfer in great depth. Moreover, it is a result of immense real-world consequence. Publications in online communities are very heterogeneous in style, content, and acceptability standards, as our sample of datasets has shown. Establishing the portability of a cyberbullying detector trained for one community is thus very important.

\subsection{Model and Training}
Each message of a dataset is preprocessed by lowercasing and tokenizing using the NLTK library\footnote{\url{https://www.nltk.org/}}. We use sequence bucketing, a technique that consists of padding the messages to match the longest message of a batch (a set of 128 messages in our case) instead of truncating or padding the entire dataset to a fixed length, which would result in either memory waste (if there are a few very long messages) or lost information (if longer messages are truncated). We use a FastText network pre-trained on Common Crawl data featuring 300 dimensions and 2 million word vectors with subword information\footnote{\url{https://github.com/facebookresearch/fastText}} to convert the words into vector representations. These vectors are then concatenated with a 60-dimensional binary vector representing 60 common characters; each character appearing in a word is marked as a 1 in this vector. This makes the system more robust to misspellings and typos (which are very common online): a word that is a misspelling of another may be a distant vector in word embedding space, but it will be nearby in character space. This is an idea lifted from \cite{brassard2019subversive}.

There exists many deep neural network architectures trained for cyberbullying detection. For our experiments, getting state-of-the-art results is not as important as getting comparable results on all datasets that will allow us to contrast their strengths and weaknesses. We thus opted for an attention bidirectional long-term-short-term model. This model features two bidirectional LSTM layer of 128 hidden units each, followed by a scaled-dot product attention, global max and average pooling layers, and finally three linear layers the size of the concatenated pooling and attention layers. We use layer normalization on the input, after the two bidirectional LSTM layers, and after the attention layer. We perform dropout after the initial layer normalization and before the last linear layer. The activation function is the gaussian error linear units (GELUs) \cite{hendrycks_gaussian_2018}. The output of our model is a probability distribution over the positive and negative classes, to which we apply a softmax function. This architecture has been proven to give good results in previous works \cite{deep_learning_across_smp} and in practice in cyberbullying Kaggle competitions \cite{noauthor_toxic_nodate,noauthor_jigsaw_nodate}. Compared to other state-of-the-art architectures, such as BERT or transformers, it has slightly worse performance but is much faster to train and less demanding in computational resources. 

We train our model using the Fused Adam optimiser from Nvidia's Apex library\footnote{\url{https://github.com/NVIDIA/apex}} and half-precision \cite{micikevicius2017mixed} to reduce training time. We use a learning rate of 0.05, a decaying gamma of 0.6 every epoch and cross-entropy loss. We use a batch size of 128 messages and train on each dataset for 15 epochs. 

\subsection{Results and Analysis}
We present in tables \ref{tab:individual_precision} and \ref{tab:individual_recall} the precision and recall performance of our model when trained using each of our training and validation datasets and tested on each of our datasets. These two metrics are the most important ones to optimize for cyberbullying detection: an ideal cyberbullying filter will block all cyberbullying messages and no messages that are not cyberbullying. Precision is the proportion of messages classified as the positive class that actually belong to it, and recall is the proportion of positive-class messages that are detected as such. While the F1-score combines both metrics, the averaging masks the performance details. 

As expected, we find that classifiers trained on each dataset have wildly varying performances on other test datasets, and can see their precision or recall drop by as much as 0.8. However, the results are far from uniformly bad. The top-performing classifier on each test dataset (marked in bold in the tables) is not always the one trained on it, and some classifiers perform as well or even better on other datasets compared to their own test dataset. This means that some generalization of cyberbullying detection is possible. Interestingly, good candidates for generalization are not related to similar data sources: datasets A and B both come from Twitter and datasets F and H both come from Wikipedia talk pages, and while the classifier trained on dataset B does perform better on dataset A than others, the other three show no special affinity for their similar datasets. Likewise, modeling a similar behavior of cyberbullying does not guarantee generalization: datasets A and E both focus on hate speech, yet they each have very poor recall on the other's test set, meaning they cannot accurately recognize messages the other labels as hate speech. 

Looking at table \ref{tab:individual_precision}, we can see that every model achieves an average precision between 0.45 and 0.63, meaning half the messages each one labels as positive class actually belongs to the negative class. This is a direct consequence of the problem described by the word vectors in section \ref{sec:similarity} and illustrated in figure \ref{t_sne}: a lot of neutral words with similar meanings are observed in multiple datasets but in the positive class of one and the negative class of the other. When such words occur in the training set of one dataset and the test set of another, it causes the test message to be misclassified. 

Looking next at recall results, we can see that classifiers often perform worse on other datasets. This is especially true for the classifiers trained on datasets B, C, D, and E; each of them picks out less than half the positive messages on average across datasets. These are also our four smallest datasets, with less than 15,000 messages each and limited vocabulary in \ref{tbl:wordcounts}. The limited variety of messages they have been trained on, compounded by the differences between datasets shown in section \ref{sec:similarity}, means filters trained on these datasets cannot generalize to new situations. By contrast, datasets F and G achieve the highest recall scores, and are able to pick out positive-class messages in other datasets despite their differences. These are also the two largest datasets, and the two that label the largest range of cyberbullying behaviors (six each, compared to one or two in other datasets). Having seen a greater variety of cyberbullying messages has allowed them to generalize better to new datasets.


Looking at performance across test datasets, we find that dataset A is the one where systems perform the best: all models achieve their highest precision on it, and all but two models achieving better than 0.5 recall. This is because it is the only dataset imbalanced in favour of the positive class. This makes the classification task easier; classifiers can be less discriminating while still avoiding mislabeling negative-class messages in this dataset in a way they cannot in other datasets where positive-class messages are a rare exception. Every classifier achieves a significantly lower precision on every other dataset, including the one it is trained for, indicating that all classifiers routinely mislabel negative-class messages as positive class in all datasets except in dataset A where negative-class messages are just too rare. On the other hand, models achieve most of the lowest precision and recall scores on dataset E. In fact, aside from the model trained specifically on dataset E, every model has trouble with that dataset, with on average only one-third of labeled messages being actually positive-class and one-sixth of positive-class messages being identified as such. This is likely due to that dataset's narrow definition, combined with the nature of the source. Dataset E includes a clear intent to attack in the definition of its positive class, and thus has messages with off-hand racial slurs labeled as negative-class due to lack of intent but picked out as positive class by other systems. However, dataset D also requires intent in its positive class, and does not suffer from the same problem. But dataset D was collected from an ordinary social network, while dataset E was collected from Stormfront, a white-supremacist and neo-nazi community. It is not surprising that this dataset includes a lot more negative-class messages with casually aggressive and hateful language that corresponds to positive-class messages of other communities and causes a low precision. The low recall is due to the fact that many of its positive-class messages use racist imagery and idioms, such as attacking people by saying they have "African blood" in them, rather than explicit hate-speech vocabulary.

\begin{table}[htbp]
\caption{Cross-dataset precision}
\begin{center}
\begin{tabular}{|c|c|c|c|c|c|c|c|c|c|}
\hline
\ & \multicolumn{8}{|c|}{Test dataset} \\
\  & A & B & C & D & E & F & G & H \\
\hline
 A & 0.98 & 0.57 & 0.31 & 0.53 & 0.27 & 0.56 & 0.48 & 0.77 \\
 B & 0.96 & \textbf{0.77} & 0.27 & 0.35 & 0.22 & 0.44 & 0.23 & 0.43 \\
 C & \textbf{0.99} & 0.69 & \textbf{0.53} & 0.63 & \textbf{0.66} & \textbf{0.65} & 0.47 & \textbf{0.87} \\
 D & 0.97 & 0.45 & 0.19 & \textbf{0.75} & 0.19 & 0.59 & 0.38 & 0.77 \\
 E & 0.97 & 0.72 & 0.41 & 0.48 & 0.55 & 0.49 & 0.24 & 0.64 \\
 F & 0.95 & 0.51 & 0.21 & 0.51 & 0.36 & 0.51 & 0.62 & 0.77 \\
 G & 0.97 & 0.52 & 0.30 & 0.55 & 0.37 & 0.58 & \textbf{0.80} & 0.81 \\
 H & 0.98 & 0.43 & 0.26 & 0.56 & 0.23 & 0.61 & 0.68 & 0.84 \\
\hline
\end{tabular}
\label{tab:individual_precision}
\end{center}
\end{table}


\begin{table}[htbp]
\caption{Cross-dataset recall}
\begin{center}
\begin{tabular}{|c|c|c|c|c|c|c|c|c|c|}
\hline
\ & \multicolumn{8}{|c|}{Test dataset} \\
 & A & B & C & D & E & F & G & H \\
\hline
 A & \textbf{0.98} & 0.21 & 0.65 & 0.67 & 0.17 & 0.65 & 0.14 & 0.63 \\
 B & 0.53 & \textbf{0.78} & 0.25 & 0.11 & 0.02 & 0.11 & 0.04 & 0.11 \\
 C & 0.63 & 0.17 & 0.57 & 0.68 & 0.06 & 0.47 & 0.12 & 0.47 \\
 D & 0.44 & 0.10 & 0.68 & 0.70 & 0.10 & 0.48 & 0.20 & 0.53 \\
 E & 0.15 & 0.07 & 0.28 & 0.24 & \textbf{0.60} & 0.19 & 0.24 & 0.17 \\
 F & 0.93 & 0.30 & \textbf{0.84} & \textbf{0.86} & 0.35 & \textbf{0.93} & 0.49 & \textbf{0.86} \\
 G & 0.81 & 0.25 & 0.74 & 0.79 & 0.37 & 0.89 & \textbf{0.57} & 0.76 \\
 H & 0.75 & 0.15 & 0.78 & 0.82 & 0.11 & 0.79 & 0.30 & 0.77 \\
\hline
\end{tabular}
\label{tab:individual_recall}
\end{center}
\end{table}


\subsection{Performance in relation to similarity between datasets}
\begin{figure}[htbp]
\centerline{\includegraphics[width=0.5\textwidth]{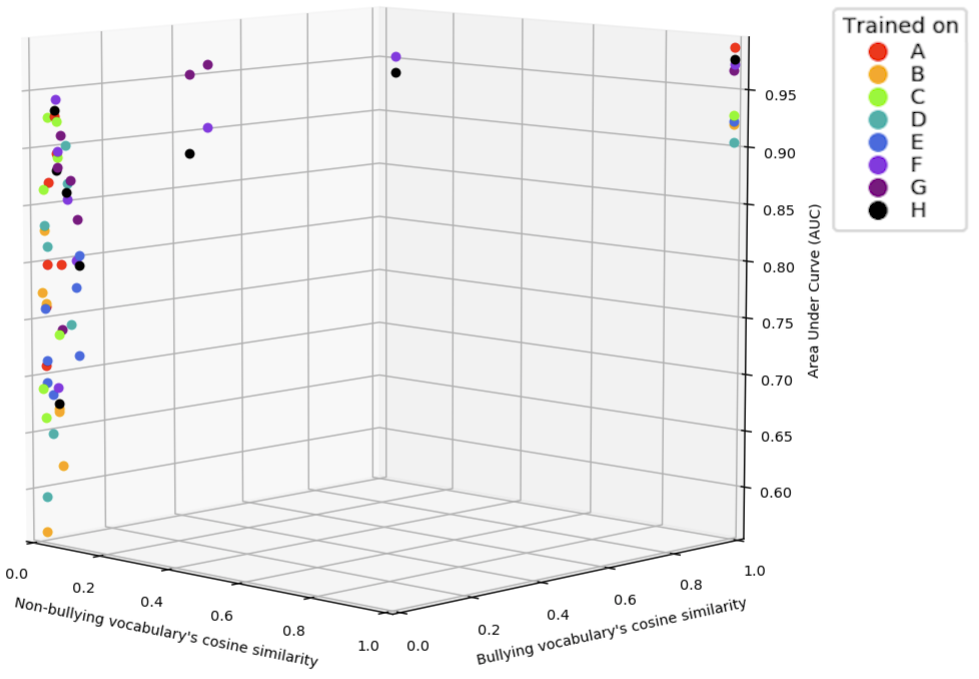}}
\caption{AUC of each test given the cosine similarity of the positive and negative datasets.}
\label{cosine_similarities_auc}
\end{figure}

In figure \ref{cosine_similarities_auc}, we consider each of our eight classifiers when tested on each of the eight datasets. The color of each of the 64 points indicates which dataset it was trained on, and its X and Y position indicates the cosine similarity between the pairs of positive and negative class datasets from table \ref{cosine_distance}. The Z axis measures the Area Under Curve (AUC) of the test, or the probability of ranking a randomly-chosen positive message higher in positive class than a negative message. Taken together, the figure plots the confidence of the classification given the dataset similarity.

We can see three clusters of points in figure \ref{cosine_similarities_auc}: a tight group near the (0,0) cosine similarity, one at (1,1) similarity, and a scattering of intermediate points. The cluster at similarity (1,1) contains the eight points where a classifier is trained and tested on the same dataset. That cluster has the best AUC points, since every classifier is optimized to its dataset. The scatter of six intermediate points are for classifiers trained on one of datasets F, G and H and tested on each of the other two. As shown in table \ref{cosine_distance}, these datasets are the most similar to each other, and this figure shows their AUC rivals that of the (1,1) cluster. Even a limited similarity between the language used in two datasets allows a classifier trained on one to perform well on the other. Finally, all other test results, 50 of our 64 points, are in the (0,0) cluster. These points cover the largest range of performances, from an AUC of 0.55 (almost random chance classification) to 0.95 (surpassing the lower half of the other two clusters). There is no coherence between datasets: the classifiers trained on datasets B and D have the worst AUC performances but also sometimes perform in the low-0.8 range, and classifiers trained on datasets F, G and H get some of the best results but also sometimes perform in the high-0.6 AUC range. Given low similarity between datasets, the performance of a classifier trained on one when applied to the other could be almost anything, and is unpredictable. Given that low similarity is the norm, this further highlights the unreliability of transferring a pre-trained cyberbullying classifier to a new community.


\section{Ensemble Experiments}\label{sec:ensemble}
If it is difficult to use a cyberbullying detector trained on one corpus to detect problem messages in another, we can wonder if it would be possible to combine a set of detectors, each trained on a different corpus, into an accurate general-domain detector. To explore this question, in this section we will compare together different ensemble model architectures built from our individual models.


\subsection{Ensemble Models}

We implemented five ensemble models that combine our existing classifiers in different ways without retraining them. \\

\subsubsection{Linear layer (LL)} In this model, the outputs of the eight individual classifiers (after the softmax function) are combined in a linear layer. This layer is trained using the same hyperparameters as the individual models, and a dropout before the linear layer ensures it does not overfit to a single model's decision. 

\subsubsection{Democratic voting (DV)} Many of the datasets are labeled independently by multiple annotators, and receive the labels chosen by the majority. We sought to replicate this logic in this ensemble model. Each of our eight classifiers casts a vote based on its classification of a message, and the winning class is simply the one with the most votes. 


\subsubsection{Sum voting (SV)}  This is a variation of the DV model that takes into account the varying levels of confidence of each model. Instead of an all-or-nothing vote for one class, each classifier votes for both the negative and positive class with the probability it assigns to each class after the softmax function. A classifier that is confident in its result will cast a strong vote for that class while one that is uncertain will cast almost equal votes for both classes and have little influence on the final decision; however, several weak votes in one class may still overrule a single strong vote in the other. 

\subsubsection{Maximum wins (MW)} This ensemble model picks the classifier with the maximum confidence in its output and assigns the message to its class. 

\subsubsection{Thresholding} In this ensemble model, if any one of the eight classifiers identifies a message as positive class with a confidence above a threshold, it is labeled as such regardless of the output of the other seven classifiers. We implemented two variations of this classifier, one with the confidence threshold at 0.5 (T0.5), which is the lowest possible confidence for a classifier to assign a message to the positive class. The other uses a threshold of 0.95 (T0.95), and if no classifier marks a message as positive with that confidence threshold the ensemble defaults to MW. The first version will thus represent extreme paranoia, where the slightest hint of cyberbullying marks a message as positive class, and the other is a paranoid version of MW.


\subsubsection{Dataset merger (DM)} As a baseline, we merged together all eight datasets and trained a new classifier on this dataset, using the training details as in section \ref{sec:generalization}.

\subsection{Results and Analysis}




Tables \ref{tab:ensemble_models_precision} and \ref{tab:ensemble_models_recall} give the precision and recall value of each ensemble technique when applied to each of our test datasets. Compared to tables \ref{tab:individual_precision} and \ref{tab:individual_recall}, we can see the ensemble models have generally better precision and worse recall. The F1-scores, which are not given due to space limitations but can be computed from the precision and recall values, of the ensemble models are comparable or better to those of the individual classifiers. This means that combining the information individual classifiers learned from training on different datasets improves the overall performances.

Looking at the performance of individual classifiers, we can see that the three voting classifiers (DV, SV and MW) have similar behaviors: they achieve some of the best precision scores and worst recall scores of all systems. This indicates that most classifiers mislabel most positive-class messages: either those messages have features of the negative class or they lack features of the positive class each classifier is trained for, or both. As a result, when the ensemble decides a message belongs in the positive class, it is usually right. However, most positive-class messages are only recognized by a minority of or low-confidence classifiers, and thus recall is low. 

The LL ensemble also combines the outputs of the eight classifiers, but using a linear layer trained to weight their individual outputs and optimize the decision. It achieves slightly worse precision but much better recall than the DV, SV and MW models. It thus behaves in the opposite manner: it catches a lot more positive-class messages, but mislabels a few more negative-class messages as well. Comparing to the results of section \ref{sec:generalization}, LL actually outperforms all but one of the individual classifiers in precision and in recall and all of them in F1-score, again confirming that there is knowledge to be gained by combining the classifiers. 

The paranoid T0.5 model achieves the top recall and lowest precision scores by wide margins. This means that most positive-class messages are labeled as such by at least one classifier, but so are a lot of negative-class messages. If used in practice, this system would create a very clean community mostly devoid of cyberbullying, but would also strongly restrict legitimate conversations. By increasing the decision threshold, T0.95 limits its positive class to messages any one classifier gets a strong signal from. This increases precision in all tests, as mildly positive messages are no longer marked in the positive class. However, the recall value decreases sharply compared to the T0.5 result, indicating that there are a lot of positive-class messages that not a single classifier can confidently recognize. Interestingly, the performance of T0.95 seems closer to LL than to T0.5.


Interestingly, the DM approach seems to give the best overall performance. It surpasses LL and T0.95 in precision and is second only to T0.5 in recall, and while it does not match DV, SV and MW in precision it is only 0.04 behind them without suffering from a drop in recall like they do. In terms of F1-score, it is also our best classifier. This means that the best way to combine the information from multiple datasets is not by combining multiple individual classifiers but by combining all datasets into a single classifier. This empirical conclusion stems also from our earlier analysis in section \ref{sec:generalization}, where we saw that diversity of language use and of cyberbullying behaviors was key to achieving good results, and similarity of language was important for generalization of the system. By combining all datasets together, the DM classifier is necessarily trained on the largest possible vocabulary and the largest set of different behaviors, and will have vocabulary similarity to all datasets. Moreover, this merger will neutralize the problem exposed in figure \ref{t_sne}, of having similar neutral-meaning words that appear only in the vocabulary of the positive class in one dataset and of the negative class in the other and thus confuse the classification. After the merger, these words appear in both classes in the unified dataset and no longer have a strong influence the classification. 

\begin{table}[htbp]
\caption{Ensemble models precision}
\begin{center}
\begin{tabular}{|c|c|c|c|c|c|c|c|c|c|}
\hline
\  &  \multicolumn{8}{|c|}{Test dataset} \\
 & A & B & C & D & E & F & G & H \\
\hline 
LL & 0.97 & 0.56 & 0.28 & 0.55 & 0.38 & 0.59 & 0.82 & 0.82  \\
 DV & \textbf{0.99} & 0.63 & \textbf{0.45} & \textbf{0.68} & 0.30 & \textbf{0.78} & 0.87 & \textbf{0.93} \\
 SV & 0.99 & 0.57 & 0.40 & 0.62 & 0.41 & 0.75 & \textbf{0.89} & 0.93 \\
 MW & 0.99 & 0.65 & 0.35 & 0.60 & \textbf{0.46} & 0.72 & 0.89 & 0.91 \\
 T0.5 & 0.93 & 0.62 & 0.14 & 0.44 & 0.31 & 0.40 & 0.33 & 0.58 \\
 T0.95 & 0.98 & 0.70 & 0.32 & 0.57 & 0.37 & 0.60 & 0.82 & 0.82 \\
 DM & 0.98 & \textbf{0.72} & 0.35 & 0.63 & 0.40 & 0.58 & 0.81 & 0.83\\
\hline
\end{tabular}
\label{tab:ensemble_models_precision}
\end{center}
\end{table}


\begin{table}[htbp]
\caption{Ensemble models recall}
\begin{center}
\begin{tabular}{|c|c|c|c|c|c|c|c|c|c|}
\hline
\ &  \multicolumn{8}{|c|}{Test dataset} \\
 & A & B & C & D & E & F & G & H \\
\hline
 LL & 0.87 & 0.30 & 0.78 & 0.82 & 0.29 & 0.91 & 0.50 & 0.78 \\
 DV & 0.69 & 0.14 & 0.66 & 0.69 & 0.07 & 0.52 & 0.11 & 0.52 \\
 SV & 0.78 & 0.17 & 0.69 & 0.74 & 0.12 & 0.63 & 0.16 & 0.62 \\
 MW & 0.87 & 0.27 & 0.65 & 0.67 & 0.11 & 0.58 & 0.163 & 0.57 \\
 T0.5 & \textbf{0.99} & \textbf{0.83} & \textbf{0.91} & \textbf{0.93} & \textbf{0.72} & \textbf{0.98} & \textbf{0.75} & \textbf{0.91} \\
 T0.95 & 0.96 & 0.47 & 0.78 & 0.78 & 0.18 & 0.84 & 0.32 & 0.75 \\
 DM & 0.98 & 0.74 & 0.65 & 0.77 & 0.38 & 0.89 & 0.54 & 0.82 \\
\hline
\end{tabular}
\label{tab:ensemble_models_recall}
\end{center}
\end{table}




\section{Conclusion}\label{sec:conclusion}

The fight against cyberbullying is a challenge of major social importance, and many datasets labeling behaviors associated with cyberbullying are available online to help train filters. In this paper, we conducted an in-depth study of the relationship between eight of these datasets and the systems that can be trained from them. First, we studied the datasets themselves, and what they tell us about cyberbullying behaviors. Next we studied the similarity in vocabulary between the datasets. We then trained deep neural network systems on each dataset and used them to study how they can be transferred from one domain to another. Finally, we studied approaches for combining the classifiers into ensemble models. 


Our paper has highlighted four major conclusions. First, there is little agreement on the definition of cyberbullying, the behaviors that comprise it, or how to measure and label them. For instance, hate speech is a recurring behavior in several datasets, but its precise definition varies so much than a classifier trained on one hate-speech dataset fails to pick it up in the others. Our second conclusion is that there is very little language in common between datasets, and what there is is often labeled in contradictory ways, which makes transferring systems difficult. In practice, this means a cyberbullying filter built for one community cannot be easily applied on another. Our third conclusion is that the condition to facilitate transferability is to have a system trained on as diverse a dataset as possible, both in terms of language use and in terms of behaviors labeled. This leads into our last conclusion, that if one wishes to combine the knowledge from different datasets in a unified system, the best way of doing this is to merge the datasets and train a single system.


However, merging multiple datasets does not solve the underlying problem, the limited vocabulary of the datasets which leads to neutral words being observed exclusively in one class and becoming false classification signals.  No additions to the dataset can completely solve that problem. Our future work will focus on an alternative solution, namely data augmentation. By replacing words in the messages by nearby words in word-embedding space, we could create a dataset that more thoroughly explores the vocabulary space. These synonyms would bridge the language gap between neutral words seen only in one class, and also create a vocabulary buffer around true positive or negative words and strengthen their predictive power. To preserve the meaning of the message in this enhancement step, we will use a context-based word embedding architecture such as the skipgram architecture \cite{mikolov2013distributed}.

\bibliographystyle{IEEEtran}
\bibliography{references.bib}
\end{document}